\documentclass[conference]{IEEEtran}
\IEEEoverridecommandlockouts
\usepackage{cite}
\usepackage{amsmath,amssymb,amsfonts}
\usepackage{algorithmic}
\usepackage{graphicx}
\usepackage{textcomp}
\usepackage{xcolor}

\begin{document}

\title{On-Device Tag Generation for Unstructured Text\\}

\makeatletter
\newcommand{\linebreakand}{%
  \end{@IEEEauthorhalign}
  \hfill\mbox{}\par
  \mbox{}\hfill\begin{@IEEEauthorhalign}
}
\makeatother

\author{\IEEEauthorblockN{Manish Chugani}
\IEEEauthorblockA{\textit{On-Device AI} \\
\textit{Samsung R\&D Institute}\\
Bangalore, India \\
m.chugani@samsung.com}
\and
\IEEEauthorblockN{Shubham Vatsal}
\IEEEauthorblockA{\textit{On-Device AI} \\
\textit{Samsung R\&D Institute}\\
Bangalore, India \\
shubham.v30@samsung.com}
\and
\IEEEauthorblockN{Gopi Ramena}
\IEEEauthorblockA{\textit{On-Device AI} \\
\textit{Samsung R\&D Institute}\\
Bangalore, India \\
gopi.ramena@samsung.com}
\linebreakand
\IEEEauthorblockN{Sukumar Moharana}
\IEEEauthorblockA{\textit{On-Device AI} \\
\textit{Samsung R\&D Institute}\\
Bangalore, India \\
msukumar@samsung.com}
\and
\IEEEauthorblockN{Naresh Purre}
\IEEEauthorblockA{\textit{On-Device AI} \\
\textit{Samsung R\&D Institute}\\
Bangalore, India \\
naresh.purre@samsung.com}
}
\maketitle

\begin{abstract}
With the overwhelming transition to smart phones, storing important information in the form of unstructured text has become habitual to users of mobile devices. From grocery lists to drafts of emails and important speeches, users store a lot of data in the form of unstructured text (for eg: in the Notes application) on their devices, leading to cluttering of data. This not only prevents users from efficient navigation in the applications but also precludes them from perceiving the relations that could be present across data in those applications. This paper proposes a novel pipeline to generate a set of tags using world knowledge based on the keywords and concepts present in unstructured textual data. These tags can then be used to summarize, categorize or search for the desired information thus enhancing user experience by allowing them to have a holistic outlook of the kind of information stored in the form of unstructured text. In the proposed system, we use an on-device (mobile phone) efficient CNN model with pruned ConceptNet resource to achieve our goal. The architecture also presents a novel ranking algorithm to extract the top n tags from any given text. 
\end{abstract}


\section{Introduction}
Data can be broadly categorized into two categories i.e. Structured Data and Unstructured Data. Structured data can be defined as text which consists of certain patterns and is highly organized. Since structured data has a defined outline and framework, machines can search and navigate through it with ease. Example of such data would be finance account number, date formats etc. Unstructured data as the name suggests, although present in abundance is very difficult to process as it does not conform to given set of rules. Examples of such data would be product reviews on e-commerce, emails etc. Structured data analysis has become a mature industry today.

Analysis of unstructured data which comprises of 80\% of enterprise data is where the actual challenge lies and the latest trend concentrates on exploiting this resource. Unstructured text contains huge amounts of unrelated and diverse information with no framework or outline for machines to be able to identify any patterns or structure in order to locate the said information. As far as unstructured text on mobile devices is concerned, it turns out that users store even more random information in the form of such text, for e.g.: passwords, otps, blog texts, to-do lists, emails, drafts for speeches, etc. This results in data of manifold nature with varied forms and lengths of text.

Our proposed system draws on knowledge of concepts, encoded in a hierarchical common-sense knowledge database known as ConceptNet\footnote{http://conceptnet.io/} to provide enhanced tag extraction capabilities. Our approach uses Deep Learning to provide abstractive extraction of concepts by using knowledge graph embeddings to extract tags from keywords while ensuring on-device efficiency by keeping the entire pipeline computationally inexpensive. Before using the knowledge graph CNN we also use Part of Speech (POS) to extract words which are nouns and proper nouns which are further fed as input to our model. Apart from these, we have also proposed a custom ranking algorithm to extract the top n tags generated from the given data.

The remaining part of the paper is organized in the following
manner: Section II talks about the related works and how
our work differs from them; Section III describes the overall
pipeline model and the techniques employed; Section IV talks
about the datasets used to either evaluate the performance of this pipeline or used as a part of this pipeline; Section V provides the experiments conducted; Section VI talks about the methods with which our pipeline has been compared to; Section VII shows the results obtained after experimentation; Section VIII talks about the applications of this pipeline in real world scenarios and Section IX finally concludes the paper and lists down some improvements which could be researched in future.

\section{Related Work}

Keyword extraction is an important task in the area of text mining. Extracting a small set of keywords from a text or document can help in various tasks in understanding the document \cite{hulth2006study}.

None of the prior works, to the best of our knowledge, have shared results on user Notes application which is one of the most prominent sources of unstructured text on-device. Additionally, our work targets predicting results with an entirely on-device pipeline. This we considered necessary so that the user privacy is maintained by not uploading his personal data on any server.

\begin{figure}
\centering
\includegraphics[width=\linewidth]{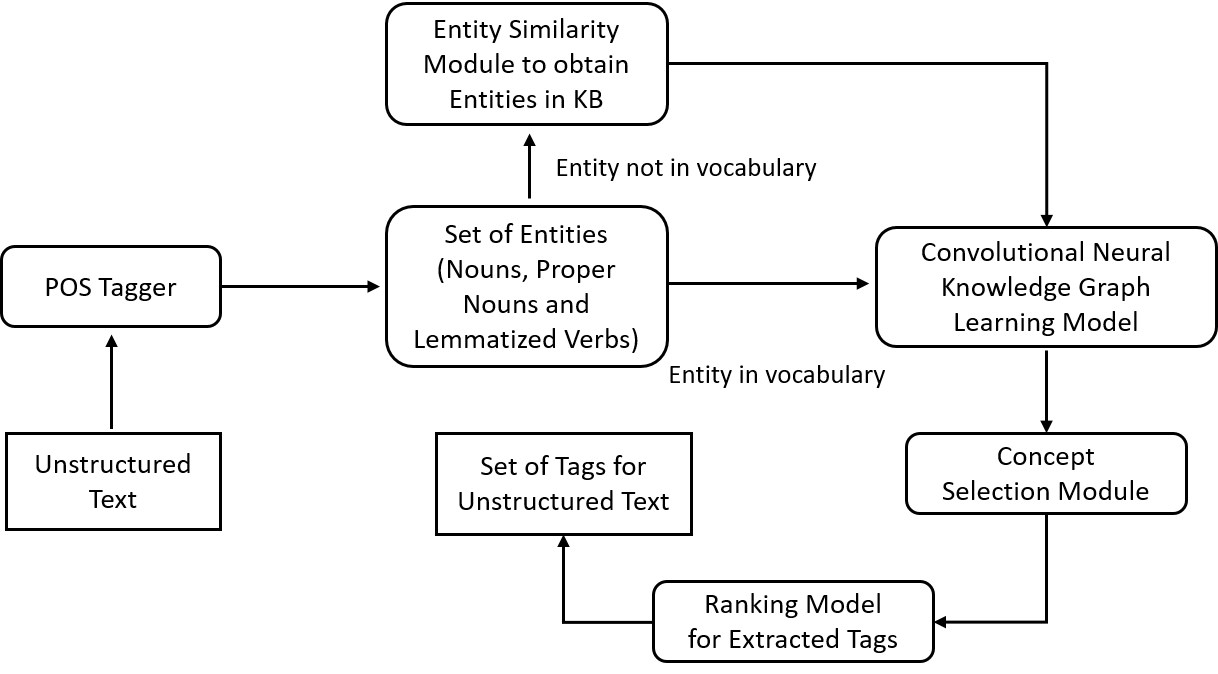}
\caption{Proposed System}
\label{fig:method}
\end{figure}

Several previous works have approached keyword extraction from short text using various statistical approaches such as TF-IDF or Bag of Words on features extracted from text. Many of these works focus in methods of selecting alternative input features. These approaches mostly rely on word frequencies and the keywords extracted are not always relevant to the user. Furnkranz et al. (1998) \cite{furnkranz1998case} uses all noun phrases matching any of a number of syntactic heuristics as features. Aizawa (2001) \cite{aizawa2001linguistic} extracts POS entities, by matching pre-defined patterns. Their representation shows a small improvement in results. In these works, it is unclear how many keywords are extracted.

Witten et al. \cite{witten2005kea} use a key phrase extraction algorithm, called KEA, based on Naive Bayes algorithm. Their algorithm learns a model for identifying extracted keywords during training, which is then applied to finding keywords from new documents. Tang et al. (2004) \cite{tang2004loss} also apply Bayesian decision theory for keyword extraction using word linkage information and thus using context features. However, these methods limit themselves to extracting keywords present in the text, and cannot extract keywords or tags based on the concept(s) present in the text.

Another interesting approach is depicted in Sahlgren and Coster (2004) \cite{sahlgren2004using} where they compute a concept-based representation from word co-occurrence data, which is combined with full-text representation. They show that this combination improved performance for their task of text categorization. Some other approaches \cite{ko2004improving} also make use of text summarization methods to find sentences containing relevant keywords. Then they use a scoring mechanism to give these sentences higher weight in their feature vectors. In this paper, we propose a fast, novel system for on-device extraction of keywords and generation of tags for unstructured text which generates tags from entities and concepts present in the text, and ranks those in order to enhance user experience.

\section{Proposed System}\label{sec:PS}

Fig. \ref{fig:method} shows the pipeline of the proposed system. As we can see, an unstructured text is sent as input to a POS Tagger from which a set of entities are extracted. Depending on whether those set of entities are present in the knowledge base or not, a set of similar entities is obtained. Finally, these set of entities are passed to a graph CNN model to extract the relevant tags in the form of keywords and concepts. Once these tags have been extracted, it is passed to a custom ranking method which reorganizes these set of tags on the basis of their priority. The in depth details of each component of the pipeline are mentioned in the coming sub-sections.

\begin{figure}
\centering
\includegraphics[width=0.9\linewidth]{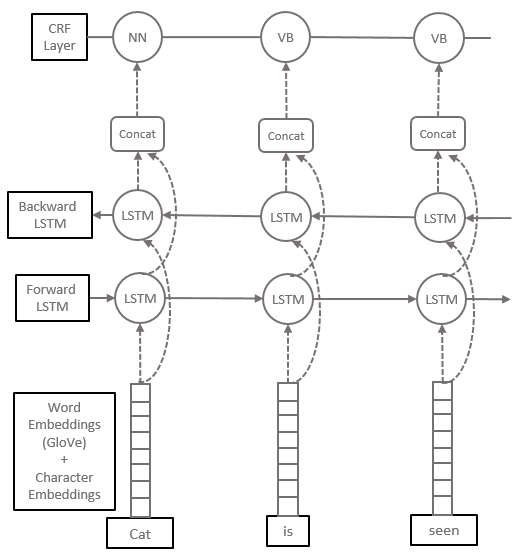}
\caption{Bi-LSTM + CRF with Character and Word (GloVe) embeddings}
\label{fig:pos2}
\end{figure}

\subsection{Part Of Speech Tagging}
For building a POS model, a model similar to Lample et al. \cite{lample2016neural} and Ma and Hovy \cite{ma2016end} is used. Firstly, a Bi-LSTM layer is trained to get character embeddings from the train data. This gives a character-based representation of each word. Next this is concatenated with standard GloVe (50 dimension vectors trained on 6 billion corpus of Wikipedia 2014 and Gigaword5) word vector representation. This gives us the contextual representation of each word. Then, a Bi-LSTM is run on each sentence represented by the above contextual representation. This final output from the model is decoded with a linear chain CRF using Viterbi algorithm. For on-device inference, the Viterbi decode algorithm is implemented in Java to be run on android devices and get the final output. The model is quantized to reduce its size and make it feasible for on-device requirements.

\begin{figure*}
\centering
\includegraphics[width=0.8\linewidth]{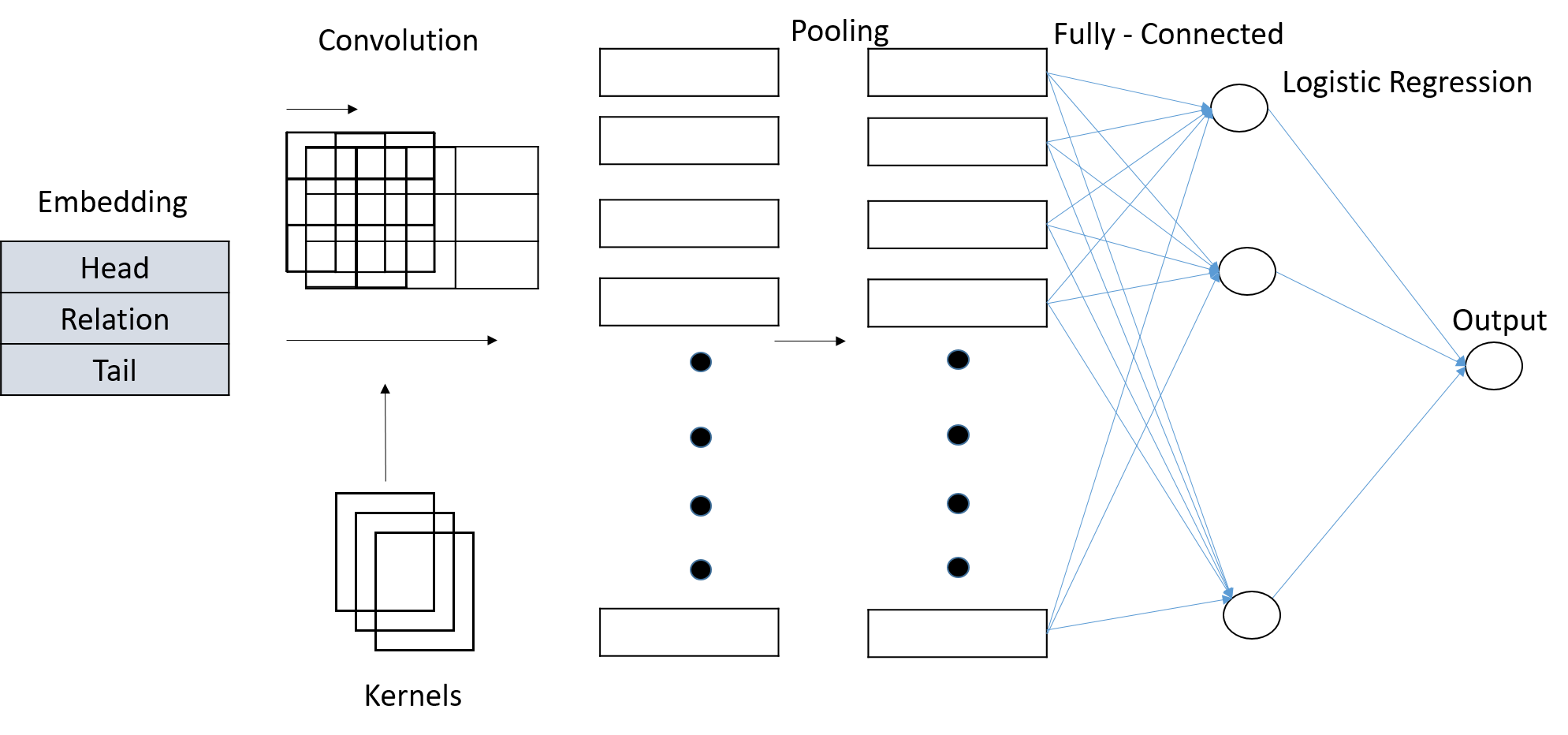}
\caption{Architecture of the Knowledge Graph CNN}
\label{fig:Architecture}
\end{figure*}

We used the tagged dataset from the CoNLL-2003 shared task for training of the above neural network. The model uses word embeddings of dimension 100, character embeddings of dimension 50 and has 100 LSTM units. The final POS model used on-device had an accuracy of 97.21\% on the test dataset. When an input text is passed to the POS model, the extracted proper nouns are added to the final set of tags. The verbs are lemmatized and passed alongside the nouns to the neural network for inferencing concepts from the commonsense based knowledge graph. Fig. \ref{fig:pos2} shows the architecture of the POS model used.

\subsection{CNN-based Knowledge Graph Learning}\label{sub:CNNKB}

Our approach uses a CNN-based knowledge graph model as explained in Feipeng Zhao et al \cite{zhao2017convolutional}. In this architecture, both embedding and CNN based score function are unknown. The model develops the entity and relation embeddings whilst also learning the knowledge graph structure using triplets (h, r, t) where h and t are the head and tail entities and r is the relationship between them.

Given any (h, r, t), the three embedding vectors are stacked over one another and 3 x 3 kernels are used for convolution over the combined matrix of size 3 x [embedding dimension]. CNNs when applied on images go through rows and columns on the image pixels but in our case they go over the locally connected structure of the head, relation and tail together. The CNN output is then passed to the max-pooling layer to get subsamples. The max-pooling filter size is set to 1 x 2 with stride as 2. Dropout is added for regularization. The dropout probability is set to 0.5 during training. The final layer of the network is a logistic-regression layer. Positive (Correct) triplets have score 1 and Negative (Incorrect) triplets have score 0. The final negative score is a tanh activation of the regression layer. The loss function is given by the formula


\begin{equation}
\sum_{(h,r,t)\epsilon S}\sum_{(h',r,t')\epsilon S'_{(h,r,t)}} 
  [\gamma + cnn(h,r,t) - cnn(h',r,t')]_{+} 
\end{equation}

where \emph{h} is the head entity, \emph{r} is the relation, \emph{t} is the tail entity, \emph{h'} is the corrupted head entity, \emph{t}' is the corrupted tail entity, \emph{S} is the set of golden triplets, \emph{S'} is the set of corrupted triplets, $\gamma$ is the margin hyperparameter of the network, \emph{cnn(h, r, t)} is the score of a golden triplet and \emph{cnn(h',r,t')} is the score of a corrupted triplet.

Mini-batch stochastic gradient descent is used as an optimizer for the loss function. Also, we require negative sampling in order to calculate the score for each positive triplet. The embedding and CNN parameters are initialized with random values. Training is fixed at a certain number of epochs based on the size of dataset used. The architecture is shown in Fig. \ref{fig:Architecture}

The training data (knowledge graph) provided to this model is filtered from the vast ConceptNet dataset as explained in Section \ref{sec:Datasets}. Our knowledge graph contains triplets of a summarizing nature and is specifically filtered for this task of generating concepts from unstructured text. Other methods use standard datasets for training and validation sets but this task required the creation of a hierarchical knowledge graph which we split in a 9:1 ratio during the model training phase.

The purpose of using a knowledge graph to generate tags is to ensure that the approach is not confined to the input text. The knowledge graph facilitates real world knowledge being applied to the extraction process to emulate human behaviour when trying to understand the same input text. Another reason for using a CNN based learning method is that this pipeline was designed for on-device inference where such models are feasible and efficient. 



\subsection{Entity Similarity Module}\label{sub:Similar}

Due to the architecture being deployed on-device, the constraints of model size and inference time are strict. This results in restrictions on how deep the CNN architecture can be since heavy model sizes prevent on-device deployment. This necessitated the use of GloVe embeddings to find similar words to entities outside the knowledge graph in order to be able to incorporate such entities. If an entity outside the vocabulary of the knowledge graph is encountered, we extract words similar to the entity in question using cosine similarity.

Table \ref{tab:Metrics} in Section \ref{sub:Model Parameters} shows the various on-device metrics of the models that have been experimented with and establishes the need for this alternative approach to incorporating large knowledge graphs on-device.




\subsection{Concept Selection Module}\label{sub:Selection}
When a word from a given text is passed through Concept-
Net, it gives a number of concepts corresponding to
that word. For example when we pass the word car through ConceptNet,
we get concepts such as \emph{artifact}, \emph{tool}, \emph{vehicle}, \emph{item}, \emph{machine}.
Most of these concepts are generally irrelevant with respect
to the general context of the text. Hence, in order to choose
the most appropriate concept, we calculate context factor. If
$c_i$ represents a concept from set of extracted concepts for a
word, $w_j$ represents an output of POS, \emph{tf($w_j$)} represents term
frequency of word $w_j$ and \emph{N} represents the length of text, then
context factor \emph{$L_{contx}(c_i)$} can be defined as

\begin{equation}
L_{contx}( c_{i}) =\frac{1}{N} \ \sum _{j} tf( w_{j}) \ *\ CosineSim( w_{j} ,c_{i})
\end{equation}

where \emph{CosineSim} is cosine similarity between word $w_j$ and
concept $c_i$ is calulcated using GloVe embeddings.

We choose the concept $c_i$ with maximum \emph{$L_{contx}$} value
as the most appropriate concept for a given word $w_i$. This
context factor helps us in analysing the general context of the
entire text while selecting a concept for a word. For example,
consider the text “Typically, the driver is responsible for all
damage to the car during the tenure of the lease, even if they
are not at fault.” In this text, \emph{$L_{contx}$(vehicle)} is maximum with
value 0.70 for the word car. But for the sentence “Machine was a very popular term in my family but car was the first machine that actually caught my imagination and I can safely say that it is my favorite machine till date.”, \emph{$L_{contx}$(machine)} is maximum with value 0.55 and hence becomes the extracted concept for the word car.



\subsection{Ranking Of Tags}

There can be a scenario that for considerably long unstructured text, we may end up extracting large number of tags, say up to 20-30 tags which can in turn prove to be another form of clutter for the user. Hence, in order to enhance user experience it is utterly important to rank and select only a handful of extracted tags for any given text. In this section, we present a custom ranking algorithm and later we also present evidence in the form of results obtained on various datasets as a justification for the hypothesis on which this algorithm is based on.

The hypothesis on which the algorithm is based is that if the tag generating word is found in the vicinity of a large number of other tag generating words for an input text, it will be given a higher priority while ranking the tags. A tag generating word is simply a word from which a tag is extracted. On the basis of our hypothesis, we calculate a ranking factor RF given by the equation

\begin{equation}
RF_{t_{j}}( t_{i}) =\sum _{j}\frac{C( w_{i} ,w_{j})}{F( w_{i}) *F( w_{j}) *\alpha }
\end{equation}

where \emph{C($w_i, w_j$)} is the co occurrence of words from which tags $t_i$ and $t_j$ have been extracted for each sentence, \emph{F($w_i$)} and \emph{F($w_j$)} are the frequencies of words $w_i$ and $w_j$ in the unstructured text and $\alpha$ is the average number of words occurring in the unstructured text between $w_i$ and $w_j$ plus 1. $RF_{t_j}$\emph{($t_i$)} is the ranking factor of tag $t_i$ with respect to tag $t_j$ . The tags are then ranked in descending order of \emph{RF} values.

\begin{table}
\centering
\caption{Table for 5 $\times$ 5 Ranking matrix}
\resizebox{\columnwidth}{!}{\begin{tabular}{|c|c|c|c|c|c|}
\hline
\textbf{} & change & vehicle  & contract & responsibility & payment  \\ \hline
change    & 1.0    & 0.11 & 0.04  & 0.02  & 0.06 \\ \hline
vehicle       & 0.11   & 1.0  & 0.13  & 0.04  & 0.02 \\ \hline
contract     & 0.04   & 0.13 & 1.0   & 0.07  & 0.08 \\ \hline
responsibility     & 0.02   & 0.04 & 0.07  & 1.0   & 0    \\ \hline
payment       & 0.06   & 0.02 & 0.08  & 0     & 1.0  \\ \hline
\end{tabular}}
\label{tab:Ranking}
\end{table}

In our custom ranking method, the co-occurrence value in the equation is determined by calculating the number of sentences in which both words $w_i$ and $w_j$ are found. The frequency for words $w_i$ and $w_j$ are calculated by taking the complete unstructured text into consideration. Another factor $\alpha$ is present which adds extra weightage to the extracted tags. This factor accounts for the distance between the words that generate tags $t_i$ and $t_j$ from the knowledge graph embeddings. The distance measure can be defined as the number of words between the words in the unstructured text from which tags $t_i$ and $t_j$ are generated. Since, our hypothesis is based on giving highest priority to a tag which occurs in the neighborhood of most of the other tags, this factor helps in achieving the same.

Here is a small example explaining the working of this algorithm. Consider the note “Typically, the driver is responsible for all damage to the car during the tenure of the lease, even if they are not at fault. Your own insurance may apply to pay for the damage. Also, the credit-card you used to pay for the lease may have supplemental insurance for damage to the car.” After this text is passed through our pipeline, the tags extracted are \emph{responsibility}, \emph{contract}, \emph{payment}, \emph{vehicle} and \emph{change}. For visualization, we construct a 5X5 ranking matrix, calculating the relatedness of these extracted tags as shown in Table \ref{tab:Ranking}.

Finally, considering the values in the ranking matrix, the pairs $w_{i}\rightarrow t_{i}$ are ranked as 
lease$\rightarrow$contract \textgreater \ car$\rightarrow$vehicle \textgreater \ damage$\rightarrow$change \textgreater \ pay$\rightarrow$payment \textgreater \ fault$\rightarrow$responsibility. Here $w_i$ is the word in the input text and $t_i$ is the extracted tag.

\section{Datasets}\label{sec:Datasets}
The dataset used for training the Convolutional Neural Knowledge Graph Learning model is ConceptNet.

The ConceptNet knowledge graph contains triplets\emph{(h, r, t)}
from various languages with a huge variety of concepts. Due
to on-device constraints, the entire ConceptNet dataset is too
vast to be inferred from. As a result of which we created
our own pruned ConceptNet dataset. We used a set of rules
in order to finally arrive at our filtered ConceptNet The first
filter we added to select a smaller set of data was to only
select those triplets that were in the English language. Another
selection technique we used was to select relationships \emph{(r in
(h, r, t))} that were such that the head entity is a superset
or parent of the tail entity. In order to ensure that the tags
extracted from unstructured textual data are of a summarizing
nature, we added this constraint. The 4 relations we used to
extract the triplets were \emph{IsA}, \emph{DerivedFrom}, \emph{InstanceOf} and
\emph{PartOf}. Other relations in the knowledge graph that were
of a slightly less summarizing nature were ambiguous and
were dropped. The ConceptNet knowledge graph also incorporated some
DBpedia relations that were filtered out since they were not
that relevant with respect to our work. This narrows down
the dataset to a few hundred thousand triplets. But this is
still too vast to be inferred from an on-device perspective
due to the model being around 200 MB after quantization
and compression. Therefore, we decided to manually select a
smaller dataset of most commonly used and relevant concepts
from the knowledge graph. This results in a dataset of around
15K triplets which reduced the model size to 2 MB after
quantization and compression. 

Apart from the dataset used for
training the Graph-CNN, we have used open source datasets
of Amazon Reviews\footnote{https://nijianmo.github.io/amazon/index.html\#files} and Enron Emails\footnote{https://www.kaggle.com/wcukierski/enron-email-dataset} for benchmarking
our proposed system. We also used a dataset of user Notes application to evaluate the feasibility of the proposed pipeline. The
Amazon Review dataset consists of short and long texts of
user reviews on various shopping categories. The Enron Email
dataset contains emails generated by employees of the Enron
Corporation. The Notes application dataset consists of Notes of variable lengths ranging from short to-do lists to
lengthy email drafts.

\section{Experiments}\label{sec:Experiments}

\subsection{Evaluation Metric for Quality of Tags}

Since the tags extracted from our text contain mostly of
concepts which are not exact same words present in our text,
we cannot use gold standard datasets to compare our method.
Another comparison method involves annotators judging the
most appropriate tags for a given piece of text but this ends
up incorporating a bias towards the authors’ own methodology
and we clearly wanted to avoid that. Inspired by Bellaachia et al. \cite{bellaachia2012ne}, we introduce
a new way to compare the quality of tags generated by various
methods. We use volume of Google search query to get an idea
about the popularity of a tag extracted.

The rationale behind using this approach is that if a keyword
is more frequently used by the masses, it must have more
significance while representing a piece of text. On an average
our method generated 9-11 tags per test sample in the datasets
mentioned in the above section. We randomly selected 5
tags extracted by our method. We then sorted them according to their
popularity and compared their search volumes one on one with
that of 5 random tags extracted from top 10 tags generated
by the given 3 methods. For comparison purposes, we made
sure we were not comparing Proper Nouns which would be
nothing but some entity names. We use Word Tracker\footnote{https://www.wordtracker.com/} to get the volume of extracted keywords and thereafter go ahead with
comparisons.

Let $t_{correct}$ be the number of keywords for a given method
which has more popularity then keywords extracted by other
methods and $t_{extracted}$ be the total number of keywords extracted
(which in our case is 5 for each sample text), \emph{Precision}
can be defined as

\begin{equation}
Precision =\frac{t_{correct}}{t_{extracted}}
\end{equation}

The comparison results of our pipeline with respect to
methods discussed in the above section are shown in Table \ref{tab:Results}

\subsection{Evaluation Metric for Quality of Tags}

We again use the volume of Google search query for the
extracted 5 tags as a measure to rank them. If a keyword
or tag is more widely searched on the internet it’s word co-occurrence
factor on which most of the ranking algorithms
are based must be of high significance for any given piece
of text. As discussed in Bellaachia et al. \cite{bellaachia2012ne} we use Binary Preference
Measure\emph{(BPM)} for calculating rank of extracted keywords. The Binary Preference Measure or BPM can be calculated as

\begin{equation}
BPM=\frac{1}{|T|} \ \sum _{t\in T} 1-\frac{|n\ ranked\ higher\ than\ t|}{|M|}
\end{equation}

where T is the set of correct tags within the set M of
tags extracted by a method and t is a correct tag
and n is an incorrect tag.

\subsection{Model Parameters}\label{sub:Model Parameters}

\begin{table}
\centering
\caption{GRAPH CNN MODEL METRICS}
\begin{tabular}{|l|l|l|l|l|}
\hline
\begin{tabular}[c]{@{}l@{}}\textbf{No. of Entities in} \\ \textbf{Knowledge Graph} \end{tabular} & \begin{tabular}[c]{@{}l@{}}\textbf{No. of} \\ \textbf{Triplets}\end{tabular} &  \begin{tabular}[c]{@{}l@{}}\textbf{Model Size}\end{tabular} & \textbf{Parameters}                                                                                                                                \\ \hline
7077                                                                                    & 13000                                                                                                                              & 5 MB                                                                        & \begin{tabular}[c]{@{}l@{}}Number of Nodes \\ in Final Layer = \\Number of Entities \\ in Knowledge Graph\end{tabular}                        \\ \hline
7077                                                                                    & 13000                                                                                                                                                                                                                   & 2 MB                                                                        & \begin{tabular}[c]{@{}l@{}}Size of Fully \\ Connected Layer = \\ Half the size of \\ Pooling Layer\end{tabular} \\ \hline
166554                                                                                  & 50000                                                                                                                            & 188 MB                                                                      & \begin{tabular}[c]{@{}l@{}}Two Convolutional \\ Layers\end{tabular}                                        \\ \hline
\end{tabular} \\

\label{tab:Metrics}
\end{table}

Our Graph-CNN model uses Adam \cite{kingma2014adam} to optimize and
learn all the parameters. In our model, we can set the width
of convolutional kernels with different size, for simplicity we
fixed the kernel size as 3x3. When using pairwise ranking loss
to learn CNN, we fixed the margin value as 1. The learning
rate in our model is fixed as 0.001. Epoch number is set as 500
for ConceptNet dataset of 13-15K triplets. We use the negative
sampling method as explained in Section \ref{sub:CNNKB}. The batch size
of triplets for mini batch Stochastic Gradient Descent is set to
500. The embedding dimension is set to 200. The dissimilarity
distance measure used is the L1 norm. The evaluation triplet
size is set to 500. The number of filters used for Convolution
is set to 8. The dropout keep probability is set to 0.5.

\begin{table}
\centering
\caption{ENTITY SIMILARITY MODULE IMPACT METRICS}
\begin{tabular}{|l|l|l|}
\hline
\textbf{Dataset} &
\begin{tabular}[c]{@{}l@{}}\textbf{Out of Vocabulary} \\ \textbf{Entities (per Test Sample)} \end{tabular} &
\begin{tabular}[c]{@{}l@{}}\textbf{Average Length} \\ \textbf{of Each Sample} \\ \textbf{(No. of Words)} \end{tabular} \\ \hline
Amazon Reviews & 2.2 & 39 \\ \hline
Enron Emails & 3.4 & 57 \\ \hline
Notes & 2.8 & 65 \\ \hline
\end{tabular}
\label{tab:EntitySim}
\end{table}

The on-device metrics for different graph CNN models
while experimenting in terms of number of triplets are listed
in Table \ref{tab:Metrics}. Model size and vocabulary length are essential
metrics that need to be taken into consideration when deploying
the model on mobile devices. As we can clearly see
from Table \ref{tab:Metrics}, the size of graph CNN model trained with
166554 entities is around 188 MB which is not at all feasible
from on-device perspectives. This is the reason we went ahead
with lightweight model along the Entity Similarity Module.
As mentioned in Section \ref{sub:Similar}, for developing a set of similar
entities in order to deal with entities outside the knowledge
graph, we optimally chose a similarity score threshold of 0.7
based on trial and error. Table \ref{tab:EntitySim} showcases the effectiveness
of our Entity Similarity Module. It shows the average number
of entities detected outside knowledge graph across all our chosen datasets.

\section{Methods For Comparison}

We used the following 3 methods for comparison with our
proposed system:

\subsection{Topic Modelling using Latent Dirichlet Allocation}

Latent Dirichlet Allocation(LDA) \cite{blei2003latent} is a generative statistical
model used in natural language processing. In Topic Modelling, it explains topics by using unobserved clusters of words which explain reasons behind some parts of data being
similar.It is an unsupervised learning model that clusters
similar groups of observations. It posits that each document is
a mixture of a small number of topics/concepts and that each
observation’s presence is attributable to one of topics of that
specific document. For our comparisons we set the number of topics to 1
and extract the top relevant keywords representing that topic.

\subsection{Automatic Summarization using Text Rank Algorithm}

Automatic Summarization is the process of computational
reduction/shortening of data in order to create a synopsis
containing highly relevant and important information whilst
abstracting the unnecessary aspects of the larger data. For
example, finding the most informative sentences from a news
article, the most representative images from a collection of
images or even the most important frames in a video fall under
the umbrella of automatic summarization.

Text Rank(TR) \cite{mihalcea2004textrank} is an unsupervised approach to automatic
summarization of text. It is a graph-based ranking
algorithm used in natural language processing. We use the
default parameters for candidate parts of speech and case of
input text and a window size of 4.

\subsection{Rapid Automatic Keyword Extraction(RAKE)}

RAKE \cite{rose2010automatic} is a popular keyword extraction technique in
natural language processing. It involves using lists of stopwords
and phrase delimiters to extract the most relevant
keywords in textual data. Python implementation of RAKE
in the rake nltk library was used with default parameters for
comparison experiments.

Common methods such as TF-IDF (Term Frequency-Inverse Document Frequency) or Bag Of Word models have not been compared with due to the length of the input texts being relatively shorter. Generating an appropriate IDF score or vocabulary for comparison would require a substantial amount of relevant text. Therefore, taking into account the average length of input texts in our specific case, we choose to not compare with such methods.

\section{Results}

Tags are extracted by the model on a set of 1500 test
samples across 3 different datasets and the evaluation metrics
mentioned in Section \ref{sec:Datasets} are used to calculate results. The
precision and BPM of the conducted experiments are shown in
Table \ref{tab:Results} and the on-device inference times and model sizes are
shown in Table \ref{tab:ondevice}. The on-device metrics have been calculated
using Samsung’s Galaxy A51 with 4 GB RAM and a 2.7 Ghz
octa-core processor.

The results clearly show an improvement in both \emph{Precision}
and \emph{BPM} on the serving data and give a quantitative perspective
to the outcomes of our proposed approach.

Apart from these results, our proposed system demonstrates
efficiency with respect to device based computational restrictions.
Our entire pipeline’s size is restricted to just around 30 MB with
inference time being as low as 670 ms. An important thing
to note here is that the overall pipeline’s size and inference
timing is more than the sum of components mentioned in
Table \ref{tab:ondevice} because of presence of additional resources like GloVe
embeddings which are used across multiple components.




\begin{table}
\centering
\caption{Results across the three datasets}
\resizebox{\columnwidth}{!}{\begin{tabular}{|l|l|l|l|l|l|l|}
\hline
& \multicolumn{2}{|l|}{\textit{Enron Email}}  & \multicolumn{2}{|l|}{\textit{Amazon Reviews}} & \multicolumn{2}{|l|}{\textit{Notes}} \\ \hline
\textbf{Methods} & \textbf{Precision}  & \textbf{BPM} & \textbf{Precision}  & \textbf{BPM} & \textbf{Precision}  & \textbf{BPM} \\ \hline
LDA & 0.13 & 0.32 & 0.19 & 0.26 & 0.18 & 0.24 \\ \hline
TR & 0.26 & 0.43 & 0.29 & 0.41 & 0.33 & 0.44 \\ \hline
RAKE & 0.24 & 0.38 & 0.31 & 0.33 & 0.27 & 0.38 \\ \hline
Proposed System & 0.47 & 0.57 & 0.42 & 0.46 & 0.49 & 0.43 \\ \hline
\end{tabular}}
\label{tab:Results}
\end{table}

\begin{table}
\centering
\caption{On-Device Inference Times and Model Sizes}
\begin{tabular}{|l|l|l|}
\hline
\textbf{Component} & \textbf{Size (MB)}  & \textbf{Inference Time per sample (ms)} \\ \hline
POS & 8 & 60 \\ \hline
Graph CNN & 2 & 500 \\ \hline
Proposed System & 31 & 670 \\ \hline
\end{tabular}
\label{tab:ondevice}
\end{table}

\begin{figure*}
\centering
\includegraphics[width=0.95\linewidth]{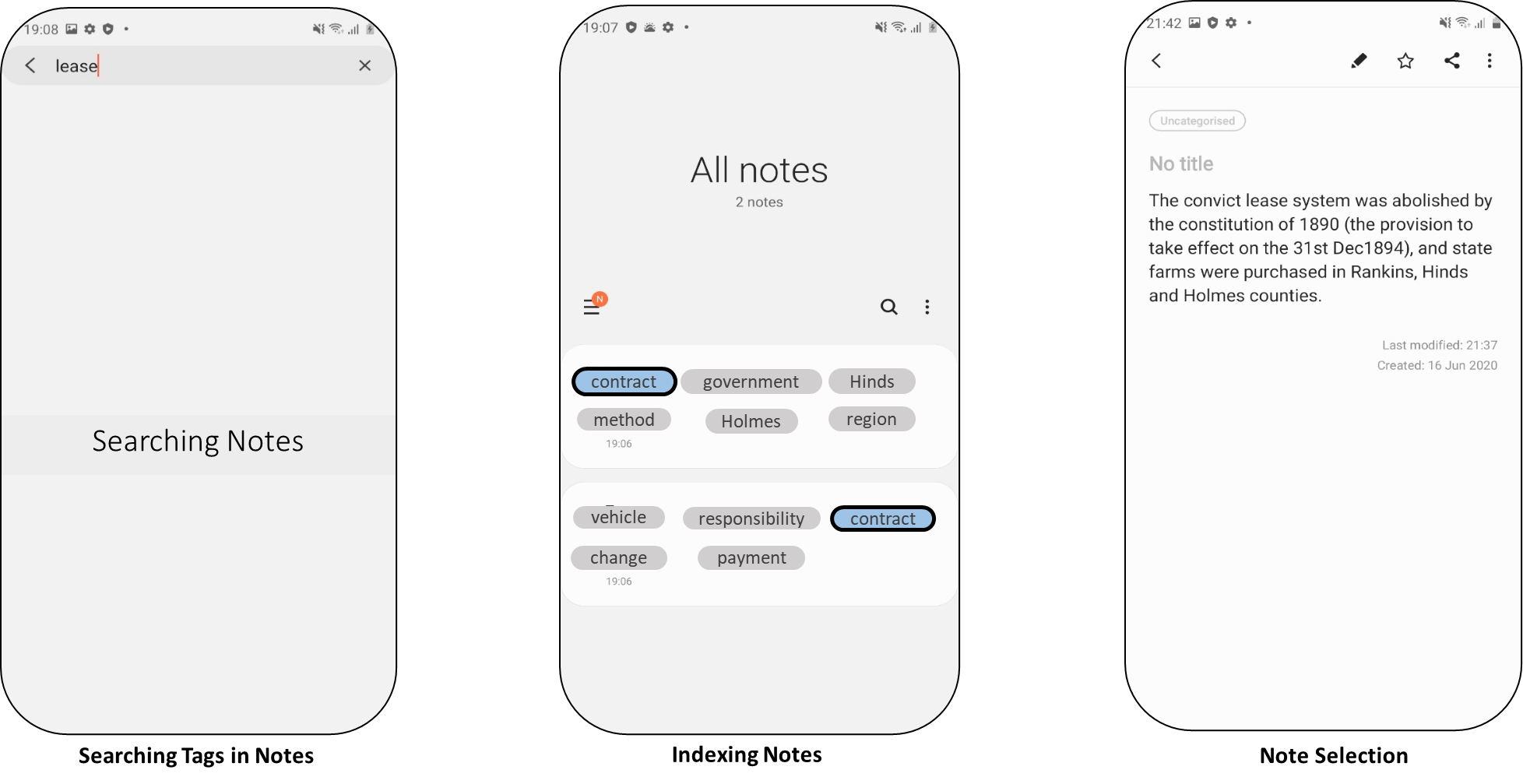}
\caption{Application Content Presentation}
\label{fig:Appflow}
\end{figure*}

\section{Applications}
\label{sec:App}

An arbitrary search for on-device Note taking applications
on the web will list down 25 to 30 such applications, thus
providing strong evidence about the utility and significance of Notes application in the modern world smartphones. From To-Do lists to email drafts, key conversations and blog snapshots, everything can be stored as a Note. Thus, Notes are a form of text that cannot be expected to have any kind of structure whatsoever and are bound to have enormous variations depending on multiple factors associated with the user. Unstructured text in Notes may or may not have punctuation marks, correct sentence formation or correct grammar.

Recently, there have been many developments in the field of Notes applications ranging from automatic detection of list type of Notes to Notes containing images but none of these new features actually address the problem of cluttering of data. In this section we show one of the ways in which this cluttering of data in Notes application can be handled using our proposed work. Fig. \ref{fig:Appflow} shows on-device screenshots which can significantly enhance user experience while navigating through Notes application. Initially in step 1, the user uses one of the querying keywords to search for the desired Note. Then in step 2, all the Notes which are indexed by the search made by the user are displayed with their set of tags extracted using our pipeline in a summarized manner. All the Notes which are indexed in step 2 are selected if there is a match between the querying keyword in step 1 and tags being displayed for each Note in step 2. Finally in step 3, user can select its desired Note out of all the indexed Notes in step 2. This is just one of the ways in which content presentation with our pipeline running in the background can be done but there can be other better ways to render the content as well.

\section{Conclusion and Future Work}

Unstructured text is a special type of text having no defined
format or pattern. Generating relevant tags in the form of concepts and keywords
from unstructured text, therefore, cannot involve the use
of contextual semantics usually associated across entire text. Thus
our proposed pipeline uses word level dynamics to extract concepts and keywords from unstructured textual data. Also, because of
disorganized nature of unstructured data, extracted tags can prove really helpful while navigating through
such text. The most popular application targeted for on-device
usage of the proposed pipeline is Notes application. With recent
developments in device based Note taking applications, our
proposed pipeline with on-device feasibility can play a vital
role in enhancing user experience as we have seen in Section \ref{sec:App}. 

One of the areas where
we can significantly improve is by analysing multiple input
data formats such as images, texts, audio etc. at the same
time. Multiple input data formats can give us a better context
and thus extracting a more subtle set of tags. Analysing these
multiple input formats would require techniques such as OCR
or speech recognition depending on the input provided by the
user.

\section{Acknowledgement}
The authors would like to thank all the users who contributed
in Notes application data collection. The authors would
like to express their gratitude towards all the reviewers who
have given constructive feedback to improve the paper.

\bibliographystyle{IEEEtran}
\bibliography{IEEEabrv,references}

\end{document}